\documentclass[11pt,a4paper]{article}
\usepackage[hyperref]{acl2017}
\usepackage{times}
\usepackage{cleveref}
\usepackage{hyperref}
\usepackage{url}
\usepackage{latexsym}
\usepackage{amsmath}
\usepackage{amssymb}
\usepackage{listings}
\usepackage{booktabs}
\usepackage{multirow}
\usepackage{float}
\usepackage{tikz}
\usepackage[framemethod=TikZ]{mdframed}
\usetikzlibrary{shapes,arrows,positioning}

\aclfinalcopy 

\title{Making Neural QA as Simple as Possible but not Simpler}

\author{Dirk Weissenborn \\\And
        Georg Wiese \\
	    Language Technology Lab, DFKI \\
	    Alt-Moabit 91c \\
	    Berlin, Germany \\
        {\tt \{dirk.weissenborn, georg.wiese, laura.seiffe\}@dfki.de} \\\And
        Laura Seiffe}

\date{}

\begin{document}
\maketitle
\begin{abstract}
Recent development of large-scale question answering (QA) datasets triggered a substantial amount of research into end-to-end neural architectures for QA. Increasingly complex systems have been conceived without comparison to simpler neural baseline systems that would justify their complexity. In this work, we propose a simple heuristic that guides the development of neural baseline systems for the extractive QA task. We find that there are two ingredients necessary for building a high-performing neural QA system: first, the awareness of question words while processing the context and second, a composition function that goes beyond simple bag-of-words modeling, such as recurrent neural networks. Our results show that FastQA, a system that meets these two requirements, can achieve very competitive performance compared with existing models. We argue that this surprising finding puts results of previous systems and the complexity of recent QA datasets into perspective.
\end{abstract}

\section{Introduction}\label{sec:intro}

Question answering is an important end-user task at the intersection of natural language processing (NLP) and information retrieval (IR). QA systems can bridge the gap between IR-based search engines and sophisticated intelligent assistants that enable a more directed information retrieval process. Such systems aim at finding precisely the piece of information sought by the user instead of documents or snippets containing the answer. A special form of QA, namely extractive QA, deals with the extraction of a direct \textit{answer} to a \textit{question} from a given textual \textit{context}.

The creation of large-scale, extractive QA datasets \cite{Rajpurkar2016,Trischler2017,Nguyen2016} sparked research interest into the development of end-to-end neural QA systems. A typical neural architecture consists of an embedding-, encoding-, interaction- and answer layer \cite{WangJiang2017,Yu2017,Xiong2017,Seo2017,Yang2017,Wang2017}. Most such systems describe several innovations for the different layers of the architecture with a special focus on developing powerful \textit{interaction layer} that aims at modeling word-by-word interaction between question and context.

Although a variety of extractive QA systems have been proposed, there is no competitive neural baseline. Most systems were built in what we call a \textit{top-down} process that proposes a complex architecture and validates design decisions by an ablation study. Most ablation studies, however, remove only a single part of an overall complex architecture and therefore lack comparison to a reasonable neural baseline. This gap raises the question whether the complexity of current systems is justified solely by their empirical results. 

Another important observation is the fact that seemingly complex questions might be answerable by simple heuristics. Let's consider the following example:

\begin{mdframed}[roundcorner=2pt]
\begin{lstlisting}[basicstyle=\small\normalfont]
|\textit{\textbf{When} did building activity occur on St. Kazimierz Church?}| 

|Building activity occurred in numerous noble palaces and churches [...]. One of the best examples [..] are Krasinski Palace (\underline{1677-1683}), Wilanow Palace (\underline{1677-1696}) and St. Kazimierz Church (\textbf{\underline{1688-1692}})|
\end{lstlisting}
\end{mdframed}
\noindent
Although it seems that evidence synthesis of multiple sentences is necessary to fully understand the relation between the answer and the question, answering this question is easily possible by applying a simple \textit{context/type matching heuristic}. The heuristic aims at selecting answer spans that a) match the expected answer type (a time as indicated by ``When") and b) are close to important question words (``St. Kazimierz Church"). The actual answer ``1688-1692" would easily be extracted by such a heuristic.

In this work, we propose to use the aforementioned \textit{context/type matching heuristic} as a guideline to derive simple neural baseline architectures for the extractive QA task. In particular, we develop a simple neural, bag-of-words (BoW)- and a recurrent neural network (RNN) baseline, namely \textit{FastQA}. Crucially, both models do not make use of a complex interaction layer but model interaction between question and context only through computable features on the word level. FastQA's strong performance questions the necessity of additional complexity, especially in the interaction layer, which is exhibited by recently developed models. We address this question by evaluating the impact of extending FastQA with an additional interaction layer (FastQAExt) and find that it doesn't lead to systematic improvements. Finally, our contributions are the following: \textbf{i)} definition and evaluation of a BoW- and RNN-based neural QA baselines guided by a simple heuristic; \textbf{ii)} bottom-up evaluation of our FastQA system with increasing architectural complexity, revealing that the awareness of question words and the application of a RNN are enough to reach state-of-the-art results; \textbf{iii)} a complexity comparison between FastQA and more complex architectures as well as an in-depth discussion of usefulness of an interaction layer; \textbf{iv)} a qualitative analysis indicating that FastQA mostly follows our heuristic which thus constitutes a strong baseline for extractive QA.

\section{A Bag-of-Words Neural QA System}\label{sec:fast_qa}

We begin by motivating our architectures by defining our proposed context/type matching heuristic: a) the type of the answer span should correspond to the expected answer type given by the question, and b) the correct answer should further be surrounded by a context that fits the question, or, more precisely, it should be surrounded by many question words. Similar heuristics were frequently implemented explicitly in traditional QA systems, e.g., in the answer extraction step of \newcite{Moldovan1999}, however, in this work our heuristic is merely used as a guideline for the construction of neural QA systems. In the following, we denote the hidden dimensionality of the model by $n$, the question tokens by $Q=(q_1, ..., q_{L_Q})$, and the context tokens by $X=(x_1, ..., x_{L_X})$.

\subsection{Embedding}\label{sec:embedder}

The embedding layer is responsible for mapping tokens $x$ to their corresponding $n$-dimensional representation $\boldsymbol{x}$. Typically this is done by mapping each word $x$ to its corresponding word embedding $\boldsymbol{x}^w$ (\textit{lookup-embedding}) using an embedding matrix $E$, s.t. $\boldsymbol{x}^w = E x$. Another approach is to embed each word by encoding their corresponding character sequence $x^c=(c_1, ..., c_{L_X})$ with $C$, s.t. $\boldsymbol{x}^c=C(x_c)$ (\textit{char-embedding}). In this work, we use a convolutional neural network for $C$ of filter width $5$ with max-pooling over time as explored by \newcite{Seo2017}, to which we refer the reader for additional details. Both approaches are combined via concatenation, s.t. the final embedding becomes $\boldsymbol{x}=\left[ \boldsymbol{x}^w ; \boldsymbol{x}^c \right] \in \mathbb{R}^d$.

\subsection{Type Matching}
For the BoW baseline, we extract the span in the question that refers to the expected, lexical answer type (LAT) by extracting either the question word(s) (e.g., \textit{who, when, why, how, how many}, etc.) or the first noun phrase of the question after the question words ``what" or ``which" (e.g., ``what \textit{year} did...").\footnote{More complex heuristics can be employed here but for the sake of simplicity we chose a very simple approach.} This leads to a correct LAT for most questions. We encode the LAT by concatenating the embedding of the first- and last word together with the average embedding of all words within the LAT. The concatenated representations are further transformed by a fully-connected layer followed by a $\tanh$ non-linearity into $\boldsymbol{\tilde{z}} \in \mathbb{R}^n$. Note that we refer to a fully-connected layer in the following by $\operatorname{FC}$, s.t. $\operatorname{FC}(\boldsymbol{u})=W\boldsymbol{u}+\boldsymbol{b}$, $W\in \mathbb{R}^{n \times m}, \boldsymbol{b}\in\mathbb{R}^n, \boldsymbol{u}\in\mathbb{R}^m$.

We similarly encode each potential answer span $(s,e)$ in the context, i.e., all spans with a specified, maximum number of words ($10$ in this work), by concatenating the embedding of the first- and last word together with the average embedding of all words within the span. Because the surrounding context of a potential answer span can give important clues towards the type of an answer span, for instance, through nominal modifiers left of the span (e.g., ``... president \underline{obama} ...") or through an apposition right of the span (e.g., ``... \underline{obama}, president of..."), we additionally concatenate the average embeddings of the $5$ words to the left and to the right of a span, respectively. The concatenated span representation, which comprises in total five different embeddings, is further transformed by a fully-connected layer with a $\tanh$ non-linearity into $\boldsymbol{\tilde{x}}_{s,e} \in \mathbb{R}^n$.

Finally, the concatenation of the LAT representation, the span representation and their element-wise product, i.e., $[\boldsymbol{\tilde{z}};\boldsymbol{\tilde{x}}_{s,e};\boldsymbol{\tilde{z}} \odot \boldsymbol{\tilde{x}}_{s,e}]$, serve as input to a feed-forward neural network with one hidden layer which computes the type score $g_{type}(s,e)$ for each span $(s,e)$.

\subsection{Context Matching}\label{sec:features}

\begin{figure*}[th]
    \centering
    \includegraphics[width=0.8\textwidth]{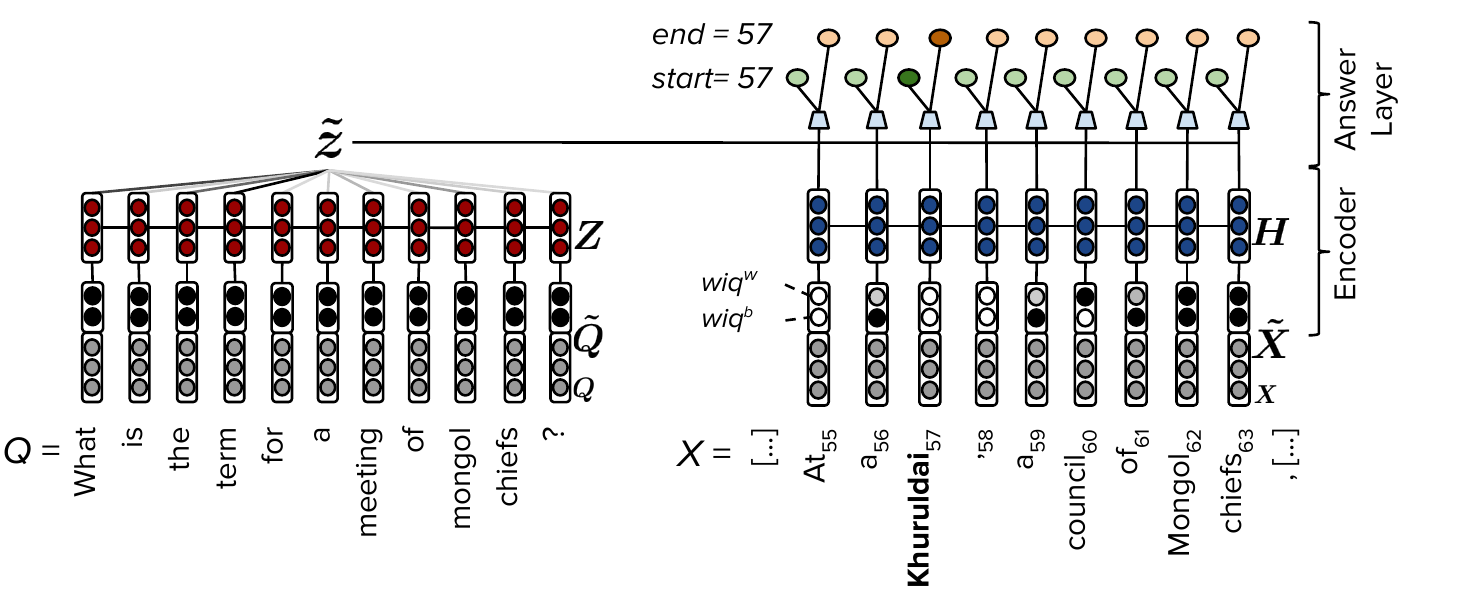}
    \caption{Illustration of FastQA system on example question from SQuAD. The two word-in-question features ($\operatorname{wiq}^b$, $\operatorname{wiq}^w$) are presented with varying degrees of activation.}
    \label{fig:fast_qa}
\end{figure*}

In order to account for the number of surrounding words of an answer span as a measure for question to answer span match (context match), we introduce two word-in-question features. They are computed for each context word $x_j$ and explained in the following

\paragraph{binary} The binary word-in-question ($\operatorname{wiq}^b$) feature is $1$ for tokens that are part of the question and else $0$. The following equation formally defines this feature where $\mathbb{I}$ denotes the indicator function:
\begin{align}
    \operatorname{wiq}^b_j = \mathbb{I}(\exists i: x_j = q_i )
\end{align}

\paragraph{weighted} The $\operatorname{wiq}^w_j$ feature for context word $x_j$ is defined in Eq.~\ref{eq:q2c}, where Eq.~\ref{eq:q2c_sim} defines a basic similarity score between $q_i$ and $x_j$ based on their word-embeddings. It is motivated on the one hand by the intuition that question tokens which rarely appear in the context are more likely to be important for answering the question, and on the other hand by the fact that question words might occur as morphological variants, synonyms or related words in the context. The latter can be captured (softly) by using word embeddings instead of the words themselves whereas the former is captured by the application of the $\operatorname{softmax}$ operation in Eq.~\ref{eq:q2c} which ensures that infrequent occurrences of words are weighted more heavily.

\begin{align}
    sim_{i,j} &= \boldsymbol{v}_{wiq} ( \boldsymbol{x}_j \odot \boldsymbol{q}_i ) \quad , \, \boldsymbol{v}_{wiq} \in \mathbb{R}^n \label{eq:q2c_sim} \\
    \operatorname{wiq}^w_j &= \sum_i \operatorname{softmax}( sim_{i,\cdot} )_j \label{eq:q2c}
\end{align}

A derivation that connects $\operatorname{wiq}^w$ with the term-frequencies (a prominent information retrieval measure) of a word in the question and the context, respectively, is provided in Appendix~\ref{sec:wiq_tf}. 

Finally, for each answer span $(s,e)$ we compute the average $\operatorname{wiq}^b$ and $\operatorname{wiq}^w$ scores of the $5$, $10$ and $20$ token-windows to the left and to the right of the respective $(s,e)$-span. This results in a total of $2$ (kinds of features)$\times 3$ (windows)$\times 2$ (left/right) $=12$ scores which are weighted by trainable scalar parameters and summed to compute the context-matching score $g_{ctxt}(s,e)$.

\subsection{Answer Span Scoring}

The final score $g$ for each span $(s,e)$ is the sum of the type- and the context matching score: $g(s,e) = g_{type}(s,e) + g_{ctxt}(s,e)$. The model is trained to minimize the $\operatorname{softmax}$-cross-entropy loss given the scores for all spans.

\section{FastQA}

Although our BoW baseline closely models our intended heuristic, it has several shortcomings. First of all, it cannot capture the compositionality of language making the detection of sensible answer spans harder. Furthermore, the semantics of a question is dramatically reduced to a BoW representation of its expected answer-type and the scalar word-in-question features. Finally, answer spans are restricted to a certain length.

To account for these shortcomings we introduce another baseline which relies on the application of a single bi-directional recurrent neural networks (BiRNN) followed by a answer layer that separates the prediction of the start and end of the answer span. \newcite{Lample2016} demonstrated that BiRNNs are powerful at recognizing named entities which makes them sensible choice for context encoding to allow for improved type matching. Context matching can similarly be achieved with a BiRNN by informing it of the locations of question tokens appearing in the context through our $\operatorname{wiq}$-features. It is important to recognize that our model should implicitly learn to capture the heuristic, but is not limited by it.

On an abstract level, our RNN-based model, called FastQA, consists of three basic layers, namely the embedding-, encoding- and answer layer. Embeddings are computed as explained in \S\ref{sec:embedder}. The other two layers are described in detail in the following. An illustration of the basic architecture is provided in Figure~\ref{fig:fast_qa}.

\subsection{Encoding}

In the following, we describe the encoding of the context which is analogous to that of the question. 

To allow for interaction between the two embeddings described in \S\ref{sec:embedder}, they are first projected jointly to a $n$-dimensional representation (Eq.~\ref{eq:emb_proj})) and further transformed by a single highway layer (Eq.~\ref{eq:highway}) similar to \newcite{Seo2017}.

\begin{align}
    \boldsymbol{x}^\prime &= P \boldsymbol{x} \quad ,\, P \in \mathbb{R}^{n \times d}  \label{eq:emb_proj} \\ 
    \boldsymbol{g}_e &= \boldsymbol{\sigma}(\operatorname{FC}(\boldsymbol{x}^\prime))\, , \,
    \boldsymbol{x}^{\prime\prime} = \tanh\left(\operatorname{FC}\left(\boldsymbol{x}^\prime\right)\right) \nonumber \\
    \tilde{\boldsymbol{x}} &= \boldsymbol{g}_e \boldsymbol{x}^\prime + (1-\boldsymbol{g}_e) \boldsymbol{x}^{\prime\prime} \label{eq:highway}    
\end{align}

Because we want the encoder to be aware of the question words we feed the binary- and the weighted \textit{word-in-question} feature of \S\ref{sec:features} in addition to the embedded context words as input. The complete input $\boldsymbol{\tilde{X}} \in \mathbb{R}^{n+2 \times L_X}$ to the encoder is therefore defined as follows:
\begin{align}
    \boldsymbol{\tilde{X}}=([\tilde{\boldsymbol{x}}_1;\operatorname{wiq}^b_1;\operatorname{wiq}^w_1], ... , [\tilde{\boldsymbol{x}}_{L_X};\operatorname{wiq}^b_{L_X};\operatorname{wiq}^w_{L_X}]) \nonumber
\end{align}

$\boldsymbol{\tilde{X}}$ is fed to a bidirectional RNN and its output is again projected to allow for interaction between the features accumulated in the forward and backward RNN (Eq.~\ref{eq:encoder}). In preliminary experiments we found LSTMs \cite{Hochreiter1997} to perform best.

\begin{align}
    \boldsymbol{H^\prime} &= \operatorname{Bi-LSTM}(\boldsymbol{\tilde{X}}) \quad,\, \boldsymbol{H^\prime} \in \mathbb{R}^{2n \times L_X} \nonumber \\
    \boldsymbol{H} &= \tanh(B \boldsymbol{H^\prime}^\top) \quad,\,  B \in \mathbb{R}^{n \times 2n} \label{eq:encoder}
\end{align}

We initialize the projection matrix $B$ with $[I_n;I_n]$, where $I_n$ denotes the $n$-dimensional identity matrix. It follows that $\boldsymbol{H}$ is the sum of the outputs from the forward- and backward-LSTM at the beginning of training.

As mentioned before, we utilize the same encoder parameters for both question and context, except the projection matrix $B$ which is not shared. However, they are initialized the same way, s.t. the context and question encoding is identical at the beginning of training. Finally, to be able to use the same encoder for both question and context we fix the two $\operatorname{wiq}$ features to $1$ for the question. 

\subsection{Answer Layer}

After encoding context $X$ to $\boldsymbol{H}= [\boldsymbol{h}_1, ..., \boldsymbol{h}_{L_X}]$ and the question $Q$ to $\boldsymbol{Z}= [\boldsymbol{z}_1, ..., \boldsymbol{z}_{L_Q}]$, we first compute a weighted, $n$-dimensional question representation $\boldsymbol{\tilde{z}}$ of $Q$ (Eq.~\ref{eq:question_representation}). Note that this representation is context-independent and as such only computed once, i.e., there is no additional word-by-word interaction between context and question.

\begin{align}
    \alpha &= \operatorname{softmax}(\boldsymbol{v}_q \boldsymbol{Z}) \quad , \, \boldsymbol{v}_q \in \mathbb{R}^n \nonumber \\
    \boldsymbol{\tilde{z}} &= \sum_i \alpha_i \boldsymbol{z}_i \label{eq:question_representation}
\end{align}

The probability distribution $p_s$ for the start location of the answer is computed by a 2-layer feed-forward neural network with a rectified-linear (ReLU) activated, hidden layer $\boldsymbol{s}_j$ as follows:

\begin{align}
    &\boldsymbol{s}_j = \operatorname{ReLU}  \left(\operatorname{FC}\left( \left[ \boldsymbol{h}_j ; \boldsymbol{\tilde{z}} ; \boldsymbol{h}_j \odot \boldsymbol{\tilde{z}} \right] \right) \right) \nonumber \\
    &p_{s}(j) \propto \exp(\boldsymbol{v}_s \boldsymbol{s}_j) \quad , \, \boldsymbol{v}_s \in \mathbb{R}^n
\end{align}

The conditional probability distribution $p_e$ for the end location conditioned on the start location $s$ is computed similarly by a feed-forward neural network with hidden layer $\boldsymbol{e}_j$ as follows:

\begin{align}
    &\boldsymbol{e}_j = \operatorname{ReLU} \left(\operatorname{FC}\left( \left[ \boldsymbol{h}_j ; \boldsymbol{h}_s  ; \boldsymbol{\tilde{z}} ; \boldsymbol{h}_j \odot \boldsymbol{\tilde{z}} ; \boldsymbol{h}_j \odot \boldsymbol{h}_s  \right] \right) \right) \nonumber \\
    &p_{e}(j|s) \propto \exp(\boldsymbol{v}_e \boldsymbol{e}_j) \quad , \, \boldsymbol{v}_e \in \mathbb{R}^n
\end{align}

The overall probability $p$ of predicting an answer span $(s,e)$ is $p(s,e) = p_s(s) \cdot p_e(e|s)$. The model is trained to minimize the cross-entropy loss of the predicted span probability $p(s,e)$.

\paragraph{Beam-search} During prediction time, we compute the answer span with the highest probability by employing beam-search using a beam-size of $k$. This means that ends for the top-$k$ starts are predicted and the span with the highest overall probability is predicted as final answer.

\section{Comparison to Prior Architectures}\label{sec:comparison}

\begin{figure}[t]
    \centering
    \includegraphics[width=0.5\textwidth]{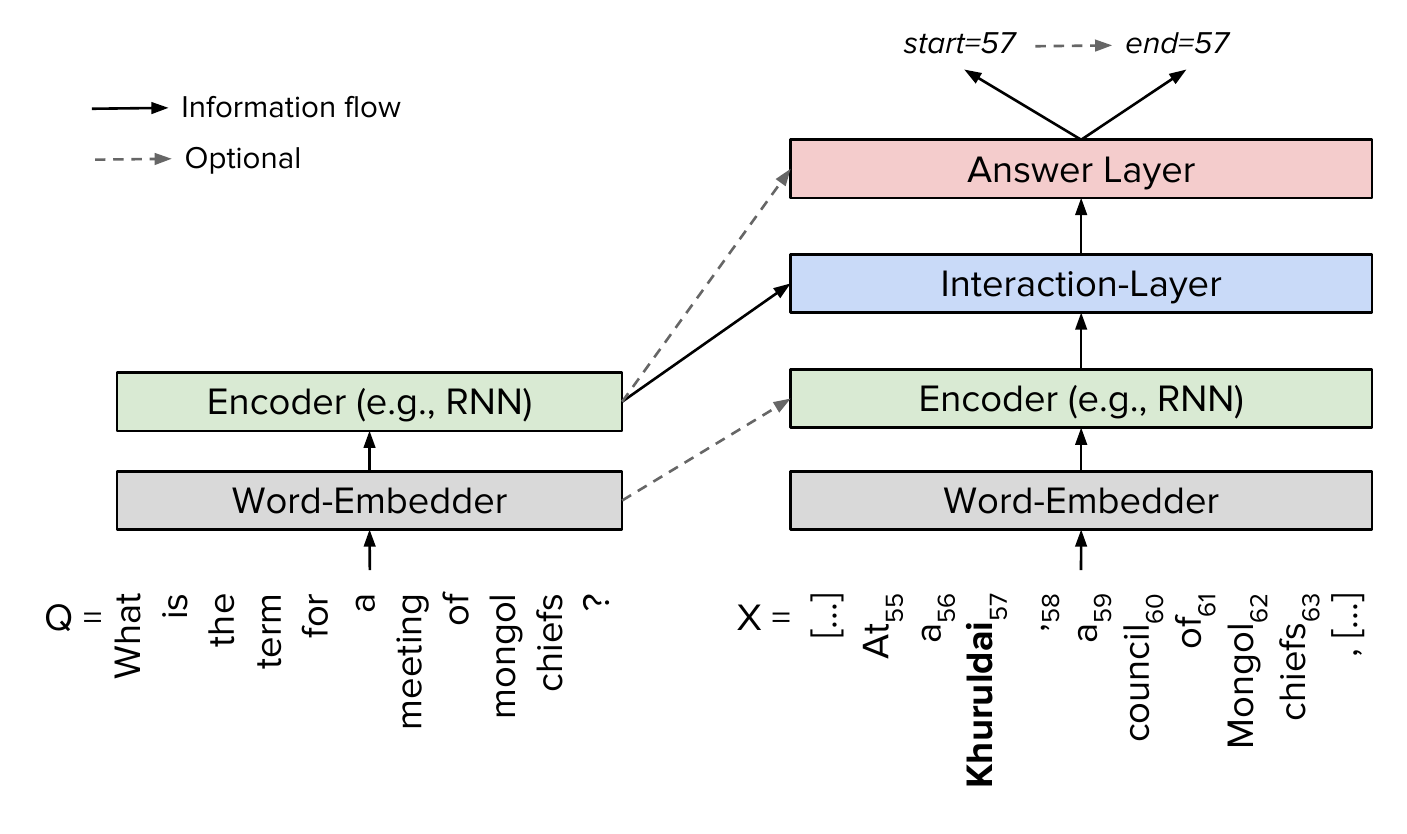}
    \caption{Illustration of the basic architecture which underlies most existing neural QA systems.}
    \label{fig:extractive_qa}
\end{figure}

Many neural architectures for extractive QA have been conceived very recently. Most of these systems can be broken down into four basic layers for which individual innovations were proposed. A high-level illustration of these systems is show in Figure~\ref{fig:extractive_qa}. In the following, we compare our system in more detail with existing models.

\paragraph{Embedder} The embedder is responsible for embedding a sequence of tokens into a sequence of $n$-dimensional states. Our proposed embedder (\S\ref{sec:embedder}) is very similar to existing ones used for example in \newcite{Seo2017,Yang2017}.

\paragraph{Encoder} Embedded tokens are further encoded by some form of composition function. A prominent type of encoder is the (bi-directional) recurrent neural network (RNN) which is also used in this work. Feeding additional word-level features similar to ours is rarely done with the exception of \newcite{Wang2017,Li2016}.

\paragraph{Interaction Layer} Most research focused on the interaction layer which is responsible for word-by-word interaction between context and question. Different ideas were explored such as attention \cite{WangJiang2017,Yu2017}, co-attention \cite{Xiong2017}, bi-directional attention flow \cite{Seo2017}, multi-perspective context matching \cite{Wang2017} or fine-grained gating \cite{Yang2017}. All of these ideas aim at enriching the encoded context with weighted states from the question and in some cases also from the context. These are gathered individually for each context state, concatenated with it and serve as input to an additional RNN. Note that this layer is omitted completely in FastQA and therefore constitutes the main simplification over previous work.

\paragraph{Answer Layer} Finally, most systems divide the prediction the start and the end by another network. Their complexity ranges from using a single fully-connected layer \cite{Seo2017,Wang2017} to employing convolutional neural networks \cite{Trischler2017} or recurrent, deep Highway-Maxout-Networks\cite{Xiong2017}. We further introduce \textit{beam-search} to extract the most likely answer span with a simple 2-layer feed-forward network.

\section{FastQA Extended}
 
To explore the necessity of the interaction layer and to be architecturally comparable to existing models we extend FastQA with an additional interaction layer (FastQAExt). In particular, we introduce \textit{representation fusion} to enable the exchange of information in between passages of the context (\textit{intra-fusion}), and between the question and the context (\textit{inter-fusion}). Representation fusion is defined as the weighted addition between a state, i.e., its $n$-dimensional representation, and its respective co-representation. For each context state its corresponding co-representation is retrieved via attention from the rest of the context (intra) or the question (inter), respectively, and ``fused" into its own representation. For the sake of brevity we describe technical details of this layer in Appendix~\ref{sec:rep_fusion}, because this extension is not the focus of this work but merely serves as a representative of the more complex architectures described in \S\ref{sec:comparison}.

\section{Experimental Setup}

We conduct experiments on the following datasets.

\paragraph{SQuAD} The Stanford Question Answering Dataset \cite{Rajpurkar2016}\footnote{\url{https://rajpurkar.github.io/SQuAD-explorer/}} comprises over $100k$ questions about paragraphs of $536$ Wikipedia articles.

\paragraph{NewsQA} The NewsQA dataset \cite{Trischler2017}\footnote{\url{https://datasets.maluuba.com/NewsQA/}} contains $100k$ answerable questions from a total of $120k$ questions. The dataset is built from CNN news stories that were originally collected by \newcite{Hermann2015}.

Performance on the SQuAD and NewsQA datasets is measured in terms of \textit{exact match} (accuracy) and a mean, per answer token-based \textit{F1} measure which was originally proposed by \newcite{Rajpurkar2016} to also account for partial matches.

\subsection{Implementation Details}

\paragraph{BoW Model}

The BoW model is trained on spans up to length 10 to keep the computation tractable. This leads to an upper bound of about $95\%$ accuracy on SQuAD and $87\%$ on NewsQA. As pre-processing steps we lowercase all inputs and tokenize it using spacy\footnote{\url{http://spacy.io}}. The binary word in question feature is computed on lemmas provided by spacy and restricted to alphanumeric words that are not stopwords.
Throughout all experiments we use a hidden dimensionality of $n=150$, dropout at the input embeddings with the same mask for all words \cite{Gal2015} and a rate of $0.2$ and $300$-dimensional fixed word-embeddings from \verb|Glove| \cite{Pennington2014}. We employed ADAM \cite{Kingma2015} for optimization with an initial learning-rate of $10^{-3}$ which was halved whenever the $F1$ measure on the development set dropped between epochs. We used mini-batches of size $32$.

\paragraph{FastQA}

The pre-processing of FastQA is slightly simpler than that of the BoW model. We tokenize the input on whitespaces (exclusive) and non-alphanumeric characters (inclusive). The binary word in question feature is computed on the words as they appear in context.
Throughout all experiments we use a hidden dimensionality of $n=300$, variational dropout at the input embeddings with the same mask for all words \cite{Gal2015} and a rate of $0.5$ and $300$-dimensional fixed word-embeddings from \verb|Glove| \cite{Pennington2014}. We employed ADAM \cite{Kingma2015} for optimization with an initial learning-rate of $10^{-3}$ which was halved whenever the $F1$ measure on the development set dropped between checkpoints. Checkpoints occurred after every $1000$ mini-batches each containing $64$ examples.

\paragraph{Cutting Context Length}

Because NewsQA contains examples with very large contexts (up to more than 1500 tokens) we cut contexts larger than $400$ tokens in order to efficiently train our models. We ensure that at least one, but at best all answers are still present in the remaining $400$ tokens. Note that this restriction is only employed during training.

\section{Results}

\subsection{Model Component Analysis}

\begin{table}[h]
    \centering
    \small
    \begin{tabular}{l c c}
        \toprule
        \textbf{Model} & \multicolumn{2}{c}{\textbf{Dev}} \\
        & F1 & Exact \\
        \midrule
        Logistic Regression$^1$ & $51.0$ & $40.0$  \\
        \midrule
        Neural BoW Baseline & $56.2$ & $43.8$ \\
        \midrule
        BiLSTM & $58.2$ & $48.7$ \\
        BiLSTM + $\operatorname{wiq}^b$ & $71.8$ & $62.3$ \\
        BiLSTM + $\operatorname{wiq}^w$ & $73.8$ & $64.3$ \\
        BiLSTM + $\operatorname{wiq}^{b+w}$ (FastQA$^*$) & $74.9$ & $65.5$ \\
        \midrule
        FastQA$^*$ + intrafusion & $76.2$ & $67.2$ \\
        FastQA$^*$ + intra + inter (FastQAExt$^*$) & $77.5$ & $68.4$ \\
        \midrule
        FastQA$^*$ + char-emb. (FastQA) & $76.3$ & $67.6$ \\
        FastQAExt$^*$ + char-emb. (FastQAExt) & $78.3$ & $69.9$ \\
        \midrule
        FastQA w/ beam-size 5 & $76.3$ & $67.8$ \\
        FastQAExt w/ beam-size 5 & $\mathbf{78.5}$ & $\mathbf{70.3}$ \\
        \bottomrule
    \end{tabular}
    \caption{SQuAD results on development set for increasingly complex architectures. $^1$\newcite{Rajpurkar2016}}
    \label{tab:squad_dev_results}
\end{table}

Table~\ref{tab:squad_dev_results} shows the individual contributions of each model component that was incrementally added to a plain BiLSTM model without features, character embeddings and beam-search. We see that the most crucial performance boost stems from the introduction of either one of our features ($\approx15\%$ F1). However, all other extensions also achieve notable improvements typically between $1$ and $2\%$ F1. Beam-search slightly improves results which shows that the most probable start is not necessarily the start of the best answer span.

In general, these results are interesting in many ways. For instance, it is surprising that a simple binary feature like $\operatorname{wiq}^b$ can have such a dramatic effect on the overall performance. We believe that the reason for this is the fact that an encoder without any knowledge of the actual question has to account for every possible question that might be asked, i.e., it has to keep track of the entire context around each token in its recurrent state. An informed encoder, on the other hand, can selectively keep track of question related information. It can further abstract over concrete entities to their respective types because it is rarely the case that many entities of the same type occur in the question. For example, if a person is mentioned in the question the context encoder only needs to remember that the ``question-person" was mentioned but not the concrete name of the person.

Another interesting finding is the fact that additional character based embeddings have a notable effect on the overall performance which was already observed by \newcite{Seo2017,Yu2017}. We see further improvements when employing representation fusion to allow for more interaction. This shows that a more sophisticated interaction layer can help. However, the differences are not substantial, indicating that this extension does not offer any systematic advantage.

\subsection{Comparing to State-of-the-Art}\label{sec:emp_results}

\begin{table}[t]
    \centering
    \small
        \begin{tabular}{l c c}
        \toprule
        \textbf{Model} & \multicolumn{2}{c}{\textbf{Test}} \\
        & F1 & Exact \\
        \midrule
        Logistic Regression$^1$ & $51.0$ & $40.4$  \\
        Match-LSTM$^2$ & $73.7$ & $64.7$ \\ 
        Dynamic Chunk Reader$^3$ & $71.0$ & $62.5$ \\
        Fine-grained Gating$^4$ & $73.3$ & $62.5$ \\
        Multi-Perspective Matching$^5$ & $75.1$ & $65.5$ \\
        Dynamic Coattention Networks$^6$ & $75.9$ & $66.2$ \\
        Bidirectional Attention Flow$^7$ & $77.3$ & $68.0$ \\
        r-net$^8$ & $77.9$ & $69.5$ \\
        \midrule
        FastQA w/ beam-size $k=5$ & $77.1$ & $68.4$ \\
        FastQAExt $k=5$ & $\mathbf{78.9}$ & $\mathbf{70.8}$ \\
        \bottomrule
    \end{tabular}
    \caption{Official SQuAD leaderboard of single-model systems on test set from 2016/12/29, the date of submitting our model. $^1$\newcite{Rajpurkar2016}, $^2$\newcite{WangJiang2017}, $^3$\newcite{Yu2017}, $^4$\newcite{Yang2017}, $^5$\newcite{Wang2017}, $^6$\newcite{Xiong2017}, $^7$\newcite{Seo2017}, $^8$ not published. Note that systems are regularly uploaded and improved on SQuAD.}
    \label{tab:squad_test_results}
\end{table}

\begin{table}[t]
    \centering
    \small
    \begin{tabular}{l c c c c}
        \toprule
        \textbf{Model} & \multicolumn{2}{c}{\textbf{Dev}} & \multicolumn{2}{c}{\textbf{Test}} \\
        & F1 & Exact & F1 & Exact \\
        \midrule
        Match-LSTM$^1$ & $48.9$ & $35.2$ & $48.0$ & $33.4$ \\
        BARB$^2$ & $49.6$ & $36.1$ & $48.3$ & $34.1$ \\
        \midrule
        Neural BoW Baseline & $37.6$ & $25.8$ & $36.6$ & $24.1$ \\
        FastQA $k=5$ & $\boldsymbol{56.4}$ & $\boldsymbol{43.7}$ & $55.7$ & $41.9$ \\
        FastQAExt $k=5$ & $56.1$ & $\boldsymbol{43.7}$ & $\boldsymbol{56.1}$ & $\boldsymbol{42.8}$ \\
        \bottomrule
    \end{tabular}
    \caption{Results on the NewsQA dataset. $^1$\newcite{WangJiang2017} was re-implemented by $^2$\newcite{Trischler2017}.}
    \label{tab:news_qa_results}
\end{table}

Our neural BoW baseline achieves good results on both datasets (Tables~\ref{tab:news_qa_results} and \ref{tab:squad_dev_results})\footnote{We did not evaluate the BoW baseline on the SQuAD test set because it requires submitting the model to \newcite{Rajpurkar2016} and we find that comparisons on NewsQA and the SQuAD development set give us enough insights.}. For instance, it outperforms a feature rich logistic-regression baseline on the SQuAD development set (Table~\ref{tab:squad_dev_results}) and nearly reaches the BiLSTM baseline system (i.e., FastQA without character embeddings and features). It shows that more than half or more than a third of all questions in SQuAD or NewsQA, respectively, are (partially) answerable by a very simple neural BoW baseline. However, the gap to state-of-the-art systems is quite large ($\approx20\%$F1) which indicates that employing more complex composition functions than averaging, such as RNNs in FastQA, are indeed necessary to achieve good performance.

Results presented in Tables~\ref{tab:squad_test_results} and \ref{tab:news_qa_results} clearly demonstrate the strength of the FastQA system. It is very competitive to previously established state-of-the-art results on the two datasets and even improves those for NewsQA. This is quite surprising when considering the simplicity of FastQA putting existing systems and the complexity of the datasets, especially SQuAD, into perspective. Our extended version FastQAExt achieves even slightly better results outperforming all reported results prior to submitting our model on the very competitive SQuAD benchmark.

In parallel to this work \newcite{Chen2017} introduced a very similar model to FastQA, which relies on a few more hand-crafted features and a 3-layer encoder instead of a single layer in this work. These changes result in slightly better performance which is in line with the observations in this work. 

\subsection{Do we need additional interaction?}

In order to answer this question we compare FastQA, a system without a complex word-by-word interaction layer, to representative models that have an interaction layer, namely FastQAExt and the Dynamic Coattention Network (DCN, \newcite{Xiong2017}). We measured both time- and space-complexity of FastQAExt and a reimplementation of the DCN in relation to FastQA and found that FastQA is about twice as fast as the other two systems and requires $2-4\times$ less memory compared to FastQAExt and DCN, respectively\footnote{We implemented all models in TensorFlow \cite{Abadi2015}.}.

In addition, we looked for systematic advantages of FastQAExt over FastQA by comparing SQuAD examples from the development set that were answered correctly by FastQAExt and incorrectly by FastQA ($589$ FastQAExt wins) against FastQA wins ($415$). We studied the average question- and answer length as well as the question types for these two sets but could not find any systematic difference. The same observation was made when manually comparing the kind of reasoning that is needed to answer a certain question. This finding aligns with the marginal empirical improvements, especially for NewsQA, between the two systems indicating that FastQAExt seems to generalize slightly better but does not offer a particular, systematic advantage. Therefore, we argue that the additional complexity introduced by the interaction layer is not necessarily justified by the incremental performance improvements presented in \S\ref{sec:emp_results}, especially when memory or run-time constraints exist.

\subsection{Qualitative Analysis}\label{sec:qual_ana}

Besides our empirical evaluations this section provides a qualitative error inspection of predictions for the SQuAD development dataset. We analyse $55$ errors made by the FastQA system in detail and highlight basic abilities that are missing to reach human level performance.

We found that most errors are based on a lack of either syntactic understanding or a fine-grained semantic distinction between lexemes with similar meanings. Other error types are mostly related to annotation preferences, e.g., answer is good but there is a better, more specific one, or ambiguities within the question or context. 

\begin{mdframed}[roundcorner=2pt]
\small
\begin{lstlisting}[title={\small Example FastQA errors. Predicted answers are underlined while correct answers are presented in boldface.}]

|Ex. 1: \textit{What religion did the Yuan discourage, to support Buddhism?}|

|Buddhism (especially \underline{Tibetan} Buddhism) flourished, although \textbf{Taoism} endured ... persecutions... from the Yuan government|

|Ex. 2: \textit{Kurt Debus was appointed what position for the Launch Operations Center?}|

|Launch Operations Center (LOC) ... Kurt Debus, \underline{a member of Dr. Wernher von Braun's ... team}. Debus was named the LOC's first \textbf{Director}|.

|Ex. 3: \textit{On what date was the record low temperature in Fresno?}|

|high temperature for Fresno ... set on \underline{July 8, 1905}, while the official record low ... set on \textbf{January 6, 1913}|

\end{lstlisting}

\end{mdframed}

A prominent type of mistake is a lack of fine-grained understanding of certain answer types (Ex. 1). Another error is the lack of co-reference resolution and context sensitive binding of abbreviations (Ex. 2). We also find that the model sometimes struggles to capture basic syntactic structure, especially with respect to nested sentences where important separators like punctuation and conjunctions are being ignored (Ex. 3).

A manual examination of errors reveals that about $35$ out of $55$ mistakes ($64\%$) can directly be attributed to the plain application of our heuristic. A similar analysis reveals that about $44$ out of $50$ ($88\%$) analyzed positive cases are covered by our heuristic as well. We therefore believe that our model and, wrt. empirical results, other models as well mostly learn a simple context/type matching heuristic. 

This finding is important because it reveals that an extractive QA system does not have to solve the complex reasoning types of \newcite{Chen2016} that were used to classify SQuAD instances \cite{Rajpurkar2016}, in order to achieve current state-of-the-art results.

\section{Related Work}

The creation of large scale cloze datasets such the DailyMail/CNN dataset \cite{Hermann2015} or the Children's Book Corpus \cite{Hill2016} paved the way for the construction of end-to-end neural architectures for reading comprehension. A thorough analysis by \newcite{Chen2016}, however, revealed that the DailyMail/CNN was too easy and still quite noisy. New datasets were constructed to eliminate these problems including SQuAD \cite{Rajpurkar2016}, NewsQA \cite{Trischler2017} and MsMARCO \cite{Nguyen2016}.

Previous question answering datasets such as MCTest \cite{Richardson2013} and TREC-QA \cite{Dang2007} were too small to successfully train end-to-end neural architectures such as the models discussed in \S\ref{sec:comparison} and required different approaches. Traditional statistical QA systems (e.g., \newcite{Ferrucci2012}) relied on linguistic pre-processing pipelines and extensive exploitation of external resources, such as knowledge bases for feature-engineering. Other paradigms include template matching or passage retrieval \cite{Andrenucci2005}.

\section{Conclusion}

In this work, we introduced a simple, context/type matching heuristic for extractive question answering which serves as guideline for the development of two neural baseline system. Especially FastQA, our RNN-based system turns out to be an efficient neural baseline architecture for extractive question answering. It combines two simple ingredients necessary for building a currently competitive QA system: a) the awareness of question words while processing the context and b) a composition function that goes beyond simple bag-of-words modeling. We argue that this important finding puts results of previous, more complex architectures as well as the complexity of recent QA datasets into perspective. In the future we want to extend the FastQA model to address linguistically motivated error types of \S\ref{sec:qual_ana}.

\section*{Acknowledgments}
We thank Sebastian Riedel, Philippe Thomas, Leonhard Hennig and Omer Levy for comments on an early draft of this work as well as the anonymous reviewers for their insightful comments. This research was supported by the German Federal Ministry of Education and 
Research (BMBF) through the projects ALL SIDES (01IW14002), BBDC (01IS14013E), and Software Campus (01IS12050, sub-project GeNIE).

\bibliography{acl2017}
\bibliographystyle{acl_natbib}

\clearpage

\appendix

\section{Weighted Word-in-Question to Term Frequency}\label{sec:wiq_tf}

In this section we explain the connection between the weighted word-in-question feature (\S\ref{sec:features}) defined in Eq.~\ref{eq:q2c} and the term frequency ($\operatorname{tf}$) of a word occurring in the question $Q=(q_1, ..., q_{L_Q})$ and context $X=(x_1, ..., x_{L_X})$, respectively. To facilitate this analysis, we repeat the equations at this point:

\begin{align}
    sim_{i,j} &= \boldsymbol{v}_{wiq} ( \boldsymbol{x}_j \odot \boldsymbol{q}_i ) \quad , \, \boldsymbol{v}_{wiq} \in \mathbb{R}^n \tag{\ref{eq:q2c_sim}} \\
    \operatorname{wiq}^w_j &= \sum_i \operatorname{softmax}( sim_{i,\cdot} )_j \tag{\ref{eq:q2c}}
\end{align}

Let us assume that we re-define the similarity score $sim_{i,j}$ of Eq.~\ref{eq:q2c_sim} as follows:

\begin{align}
    sim_{i,j} &= \begin{cases} 0 & \text{ if } q_i = x_j \\ -\inf & \text{ else}  \end{cases} \label{eq:new_sim}
\end{align}

Given the new (discrete) similarity score we can derive the following equation for the $\operatorname{wiq}^w$ feature for context word $x_j$. Note that we refer to the term frequency of a word $z$ in the context and question by $\operatorname{tf}(z|C)$ and $\operatorname{tf}(z|Q)$, respectively.

\begin{align*}
    \operatorname{wiq}^w_j &= \sum_i \operatorname{softmax}( sim_{i,\cdot} )_j \\
                           &= \sum_i \frac{\exp(sim_{i,j})}{\sum_{j^\prime} \exp(sim_{i,j^\prime})} \\
                           &= \sum_i \frac{\mathbb{I}(x_j = q_i)}{\sum_{j^\prime} \mathbb{I}(x_{j^\prime} = q_i)} \\
                           &= \sum_i \frac{\mathbb{I}(x_j = q_i)}{\operatorname{tf}(q_i|C)}  \\
                           &= \sum_{i,\, q_i=x_j} \frac{1}{\operatorname{tf}(q_i|C)} = \frac{\operatorname{tf}(x_j|Q)}{\operatorname{tf}(x_j|C)}
\end{align*}

Our derived formula shows that $\operatorname{wiq}^w$ of context word $x_j$ would become a simple combination of the term frequencies of $x_j$ within the context and question if our similarity score is redefined as in Eq.~\ref{eq:new_sim}. Note that this holds true for any finite value chosen in Eq.~\ref{eq:new_sim} and not just $0$.

\section{Representation Fusion}\label{sec:rep_fusion}
\subsection{Intra-Fusion}
It is well known that recurrent neural networks have a limited ability to model long-term dependencies. This limitation is mainly due to the information bottleneck that is posed by the fixed size of the internal RNN state. Hence, it is hard for our proposed baseline model to answer questions that require synthesizing evidence from different text passages. Such passages are typically connected via co-referent entities or events. Consider the following example from the NewsQA dataset \cite{Trischler2017}:

\begin{mdframed}[roundcorner=2pt]
\begin{lstlisting}[basicstyle=\small\normalfont]
|\textit{Where is Brittanee Drexel from?}|

|The \underline{mother} of a 17-year-old \textbf{Rochester,} \textbf{New York} high school student ... says she did not give her daughter permission to go on the trip. \textit{Brittanee} Marie \textit{Drexel}'s \underline{mom} says|
\end{lstlisting}
\end{mdframed}
\noindent
To correctly answer this question the representations of \textit{Rochester, New York} should contain the information that it refers to \textit{Brittanee Drexel}. This connection can, for example, be established through the mention of \textit{mother} and its co-referent mention \textit{mom}. Fusing information from the context representation $\boldsymbol{h}_\text{mom}$ into $\boldsymbol{h}_\text{mother}$ allows crucial information about the mentioning of \textit{Brittanee Drexel} to flow close to the correct answer. We enable the model to find co-referring mentions via a normalized similarity measure $\beta$ (Eq.~\ref{eq:intra_sim}). For each context state we retrieve its \textit{co-state} using $\beta$ (Eq.~\ref{eq:intra_co_state}) and finally fuse the representations of each state with their respective co-state representations via a gated addition (Eq.~\ref{eq:intra_fusion}). We call this procedure \textit{associative representation fusion}. 

\begin{align}
    \hat{\beta}_{j,k} &= \mathbb{I}(j \neq k) \, \boldsymbol{v}_{\beta} \, ( \boldsymbol{h}_j \odot \boldsymbol{h}_k ) \nonumber \\
    \beta_j &= \operatorname{softmax}(\hat{\beta}_{j,\cdot}) \label{eq:intra_sim} \\
    \boldsymbol{h}_j^{co} &= \sum_k \beta_{j,k}  \boldsymbol{h}_k  \label{eq:intra_co_state} \\    
    \boldsymbol{h}_j^\ast &= \operatorname{FUSE}(\boldsymbol{h}_j, \boldsymbol{h}_j^{co}) \nonumber \\
    &= \boldsymbol{g}_\beta \boldsymbol{h}_j + (1-\boldsymbol{g}_\beta)\boldsymbol{h}_j^{co}
    \label{eq:intra_fusion} \\
    \boldsymbol{g}_\beta &= \boldsymbol{\sigma}(\operatorname{FC}([\boldsymbol{h}_j;\boldsymbol{h}_j^{co}])) \nonumber
\end{align}

We initialize $\boldsymbol{v}_{\beta}$ with $\boldsymbol{1}$, s.t. $\hat{\beta}_{j,k}$ is initially identical to the dot-product between hidden states.

We further introduce \textit{recurrent representation fusion} to sequentially propagate information gathered by associative fusion between neighbouring tokens, e.g., between the representation of \textit{mother} containing additional information about \textit{Brittanee Drexel} and those representations of \textit{Rochester, New York}. This is achieved via a recurrent backward- (Eq.~\ref{eq:backward_fusion}) followed by a recurrent forward fusion (Eq.~\ref{eq:forward_fusion}) 

\begin{align}
    \boldsymbol{\tilde{h}}^{bw}_j &= \operatorname{FUSE}(\boldsymbol{h}^\ast_j, \boldsymbol{\tilde{h}}^{bw}_{j+1}) \label{eq:backward_fusion} \\
    \boldsymbol{\tilde{h}}_j &= \operatorname{FUSE}(\boldsymbol{\tilde{h}}^{bw}_j, \boldsymbol{\tilde{h}}_{j-1})
    \label{eq:forward_fusion}
\end{align}

Note, that during representation fusion no new features are computed but simply combined with each other. 

\subsection{Inter-Fusion}

The representation fusion between question and context states is very similar to the intra-fusion procedure. It is applied on top of the context representations after intra-fusion has been employed. Associative fusion is performed via attention weights $\gamma$ (Eq.~\ref{eq:inter_sim}) between question states $\boldsymbol{z}_i$ and context states $\boldsymbol{\tilde{h}}_j)$. The co-state is computed for each context state via $\gamma$ (Eq.~\ref{eq:inter_co_state}).

\begin{align}
    \hat{\gamma}_{i,j} &= \boldsymbol{v}_{\gamma} \, ( \boldsymbol{z}_i \odot \boldsymbol{\tilde{h}}_j ) \nonumber \\
    \gamma_i &= \operatorname{softmax}(\hat{\gamma}_{i,\cdot}) \label{eq:inter_sim} \\
    \boldsymbol{\tilde{h}}_j^{co} &= \sum_i \gamma_{i,j}  \boldsymbol{z}_i  \label{eq:inter_co_state}
\end{align}

Note, because the $\operatorname{softmax}$ normalization is applied over the all context tokens for each question word, $\gamma_i$ will be close to zero for most context positions and therefore, its co-state will be close to a zero-vector. Therefore, only question related context states will receive a non-empty co-state. The rest of inter-fusion follows the same procedure as for intra-fusion and the resulting context representations serve as input to the answer layer.

In contrast to existing interaction layers which typically combine representations retrieved through attention by concatenation and feed them as input to an additional RNN (e.g., LSTM), our approach can be considered a more light-weight version of interaction.

\section{Generative Question Answering}

Although our system was designed to answer question by extracting answers from a given context it can also be employed for generative question answering datasets such as the Microsoft Machine Reading Comprehension (MsMARCO, \newcite{Nguyen2016})\footnote{\url{http://www.msmarco.org/}}. MsMARCO contains $100k$ real world queries from the Bing search engine and human generated answers that are based on relevant web documents. Because we focus on extractive question answering in this work, we limit the queries for training to queries whose answers are directly extractable from the given web documents. We found that $67.2\%$ of all queries fall into this category. Evaluation, however, is performed on the entire development and test set, respectively, which makes it impossible to answer the subset of \textit{Yes/No} questions ($\approx7\%$) properly. For the sake of simplicity, we concatenate all given paragraphs and treat them as a single document. Since most queries in MsMARCO are lower-cased we also lower-cased the context. The official scoring measure of MsMARCO for generative models is ROUGE-L and BLEU-1. Even though our model is extractive we use our extracted answers as if they were generated.

The results are shown in Table~\ref{tab:MsMARCO_results}. The strong performance of our purely extractive system on the generative MsMARCO dataset is notable. It shows that answers to Bing queries can mostly be extracted directly from web documents without the need for a more complex generative approach. Since this was only an initial experiment on generative QA using extractive QA and the methodology used for training, pre- and post-processing on this dataset for the other models, especially for \newcite{WangJiang2017}, is unclear, the comparability to the other QA systems is limited.

\begin{table}[t]
    \centering
    \small
    \begin{tabular}{l c c c c}
        \toprule
        \textbf{Model} & \multicolumn{2}{c}{\textbf{Dev}} & \multicolumn{2}{c}{\textbf{Test}} \\
        & Bleu & Rouge & Bleu & Rouge \\
        \midrule
        ReasoNet$^1$ & - & - & $14.8$ & $19.2$ \\
        Match-LSTM$^2$ & - & - & $\boldsymbol{37.3}$ & $\boldsymbol{40.7}$ \\ 
        \midrule
        FastQA $k=5$ & $34.9$ & $33.0$ & $34.0$ & $32.1$  \\
        FastQAExt $k=5$ & $\boldsymbol{35.0}$ & $\boldsymbol{34.4}$ & $33.9$ & $33.7$ \\
        \bottomrule
    \end{tabular}
    \caption{MsMARCO leaderboard of single-model  systems on  test  set  from  2017/02/07,  the date  of  submitting  our  model. $^1$\newcite{Shen2016}- extractive model trained out-of-domain on SQuAD, $^2$\newcite{WangJiang2017}.}
    \label{tab:MsMARCO_results}
\end{table}


%


\end{document}